\documentclass{article}

    \usepackage[preprint]{neurips_2020}

\usepackage[utf8]{inputenc} %
\usepackage[T1]{fontenc}    %
\usepackage{hyperref}       %
\usepackage{url}            %
\usepackage{booktabs}       %
\usepackage{amsfonts}       %
\usepackage{nicefrac}       %
\usepackage{microtype}      %

\usepackage{url}
\usepackage{graphicx}
\usepackage{caption}
\usepackage{xcolor}
\usepackage{float}
\usepackage{amsmath}
\usepackage{subfigure}
\usepackage{wrapfig}

\usepackage[mathscr]{euscript}

\newcommand{\Ebf}{\mathbf{E}}

\newcommand{\Acal}{\mathcal{A}}
\newcommand{\Bcal}{\mathcal{B}}

\newcommand{\Lcal}{\mathcal{L}}

\newcommand{\Scal}{\mathcal{S}}

\newcommand{\Pscr}{\mathscr{P}}

\title{On Catastrophic Interference in Atari 2600 Games}

\author{%
  William Fedus\thanks{Equal contribution. Correspondence to \texttt{liamfedus@google.com} } \\
  Google Brain\\
  University of Montreal (Mila)\\
   \And
   Dibya Ghosh\footnotemark[1]\\
   Google Brain \\
   \And
   John Martin \\
   Google Brain \\
   \AND
   Marc G. Bellemare \\
   Google Brain \\
   \And
   Yoshua Bengio \\
   University of Montreal (Mila) \\
   CIFAR Senior Fellow \\
   \And
   Hugo Larochelle \\
   Google Brain \\
}

\begin{document}

\maketitle

\begin{abstract}

Model-free deep reinforcement learning is sample inefficient. One hypothesis -- speculated, but not confirmed -- is that \emph{catastrophic interference} within an environment inhibits learning. We test this hypothesis through a large-scale empirical study in the Arcade Learning Environment (ALE) and, indeed, find supporting evidence. We show that interference causes performance to plateau; the network cannot train on segments beyond the plateau without degrading the policy used to reach there. By synthetically controlling for interference, we demonstrate performance boosts across architectures, learning algorithms and environments. A more refined analysis shows that learning one segment of a game often increases prediction errors elsewhere. Our study provides a clear empirical link between catastrophic interference and sample efficiency in reinforcement learning.

\end{abstract}

\section{Introduction}
Peeling back the notable successes in deep reinforcement learning (RL) reveals an enormous \emph{sample inefficiency} \citep{mnih2015human, silver2016mastering}.
Many successful algorithms acknowledge this inefficiency and
are specifically designed to churn through data as quickly as possible, for example through parallel actors and hardware accelerators \citep{espeholt2018impala, badia2020agent57}.
One hypothesis --- previously without empirical support --- is that sample inefficiency is a result of \emph{catastrophic interference} \citep{mccloskey1989catastrophic, french1999catastrophic} within an environment.
This paper examines this hypothesis through a large-scale study of catastrophic interference in the Arcade Learning Environment (ALE) \citep{bellemare2013arcade}.

That catastrophic interference can occur within a \emph{single} game is in many ways counter-intuitive. Catastrophic interference is a well-studied phenomenon in multi-task learning, where networks trained on a new task ``forget'' how to complete previously solved tasks \citep{li2017learning, lopez2017gradient}, often observed from learning curve plateaus \citep{schaul2019ray}.
However, most Atari 2600 games have been long considered to be single tasks \citep{kirkpatrick2017overcoming}.
We present evidence to the contrary and show that interference arises \emph{within} the majority of games.
This leads to non-intuitive and unexpected consequences: we find that learning in \textsc{Breakout} interferes with the agent's performance in \textsc{Breakout}.

 To test whether the performance plateaus of deep RL agents are a result of interference, we introduce an experiment that controls for its effects. We call this the \emph{Memento}\footnote{This is a reference to the eponymous psychological thriller where the protagonist suffers from amnesia and must deduce a reasonable plan with no memory of how he arrived or even his initial goal.} experiment. 
 When an agent's performance reaches a plateau, its value network is copied to define a new, Memento agent. The Memento agent begins each episode from the final position of the preceding agent -- wherever it plateaued -- and it continues to train on experience that ensues from that point. Creating two agents in this way enables us to control for interference, because each parameter set is independent and incapable of interfering by construction. When the Memento agent makes further progress (e.g. a higher score), it indicates the original agent may have been experiencing significant interference. We find that further progress is possible on most Atari games. We observe that this phenomenon holds across many different deep RL algorithms, and that it cannot not be replicated in a single agent with greater network capacity or increased training time, isolating interference as the only salient factor.

In a more fine-grained analysis of interference, we analyze how learning on one segment of a game can affect prediction errors elsewhere.
To do so, we establish the concept of an intra-game \emph{task} via game score \citep{jain2019context}.
Game score, while not a perfect demarcation of task boundaries, is an easily extracted proxy with reasonable properties.
Changes in game score directly impacts the temporal-difference (TD) error through the reward and often represent milestones such as acquiring an object, navigating a room, or defeating an enemy.
We show that training on certain tasks causes prediction errors in other tasks to change unpredictably.
In a few environments, learning one game segment induces generalization elsewhere, but in the majority, learning results in sharp prediction error increases elsewhere.

Our paper highlights the prevalence of catastrophic interference in deep reinforcement learning.
The Memento experiment demonstrates that a Rainbow agent designed to be non-interfering improves by a median of +25\% over a standard agent across the entire Arcade Learning Environment.
Our analysis of individual games further reveals signatures of learning interference  commonly associated with continual and multitask learning \citep{parisi2019continual} within games that have long been considered to be a single task.
Together, these results provide a clear link between catastrophic interference and issues of sample efficiency, performance plateaus, and exploration.

\section{Background}
Our study is set within the standard paradigm for reinforcement learning. This describes a process by which an agent learns to act optimally from rewards gathered after taking actions. At each time step $t\in\mathbb{N}$, the agent transitions from state $S_t$ to $S_{t+1}$ after taking action $A_t$ and receiving reward $R(S_t,A_t) \in \mathbb{R}$. This process is formalized as a Markov decision process \citep{putterman1994mdp} $\left<\Scal,\Acal,R,P,\gamma\right>$, whose transitions are defined over the set of states $\Scal$ and actions $\Acal$. These occur with probability $P(S_{t+1}|S_t,A_t)$, and their future rewards are discounted according to $\gamma\in[0,1)$. Value-based agents in such settings strive to estimate for all $(s,a)$ the expected sum of discounted future reward, also known as the \textit{value function} \citep{sutton1998reinforcement}:
\begin{align}\label{eq:qvalue}
    Q_\pi(s,a) &= \Ebf_{P,\pi}\left[\sum_{t=0}^{\infty}\gamma^t R(S_t,A_t) \bigg| S_0=s, A_0=a\right].
\end{align}
The value function describes the benefit of following the (possibly) stochastic policy $\pi \colon \Scal \rightarrow \Pscr(\mathbb{R})$ after taking action $a$ in state $s$.\footnote{$\Pscr(\cdot)$ denotes the set of distributions supported on some set.} In domains of large scale, the value function is often approximated with a member of a parametric function class, $Q_\theta$, such as a neural network. Parameters $\theta\in \mathbb{R}^d$ are fit online using experience samples of the form $(s,a,r,s')$. This experience is typically collected into a buffer $\Bcal$ from which batches are later drawn at random to form a stochastic estimate of the loss
\begin{align}\label{eq:dqn_loss}
    \Lcal(\theta) = \Ebf_{\mu} \left[L\left(r + \gamma \max_{a'\in\Acal} Q_{\tau}(s', a') - Q_{\theta}(s,a)\right) \right].
\end{align}
In general, the parameters used to compute the target, $\tau$, are a prior copy of those used for action selection. Here $L \colon \mathbb{R}\rightarrow \mathbb{R}$ is the agent's loss function, and $\mu \in \Pscr(\Bcal)$ is the distribution that defines its sampling strategy. Our study includes a uniform strategy, which applies to the Deep Q-Network (DQN) \citep{mnih2013playing, mnih2015human}, and a stratified strategy known as \textit{prioritized experience replay} \citep{schaul2015prioritized}, which the C51 \citep{bellemare2017distributional} and Rainbow agent \citep{hessel2018rainbow} architectures both use . All three of these agents approximate value function with a convolutional neural network, mapping images to $Q$-values, and thus belong to the class of Deep RL methods.

\section{The Memento Experiment}\label{sec: observation}
Multi-task learning and catastrophic interference research in the ALE typically assumes that each game is a task and that multi-task learning therefore corresponds to multiple games \citep{kirkpatrick2017overcoming, fernando2017pathnet} or to different game modes \citep{farebrother2018generalization}.
Instead in this work, the central hypothesis we examine is whether multi-task dynamics and interference effects cause performance plateaus and sample inefficiency within a \emph{single game}.
We test this hypothesis with the Memento experiment.

\begin{figure}[t]
    \centering
    \subfigure[Baseline Rainbow-CTS agent achieves a maximum achieved score of 6600 (5 seeds).]{\label{fig:f1}\includegraphics[width=0.45\columnwidth]{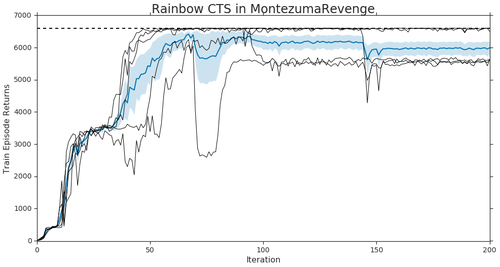}}\hfill
    \subfigure[A \emph{non-interfering} agent, whether randomly initialized (blue) or initialized with weights of the original model (orange) make further progress.]{\label{fig:f2}\includegraphics[width=0.45\columnwidth]{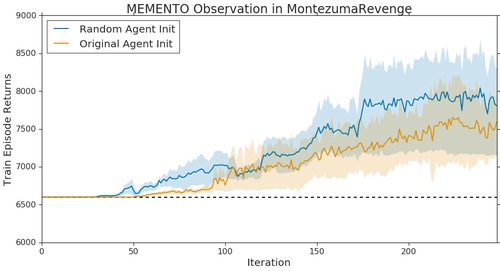}}
    \caption{The Memento experiment in \textsc{Montezuma's Revenge}. Baseline Rainbow-CTS on left plateaus at a game score of 6600 for 200M frames. On the right, both an identical, randomly initialized Rainbow CTS agent (blue) and a cloned (same initial weights) Rainbow-CTS (orange) quickly exceed plateau and result in further progress.}
    \vspace{-.5em}
\end{figure}

We illustrate the Memento experiment through \textsc{Montezuma's Revenge}, a hard exploration game initially conjectured by \citet{schaul2019ray} to have interference effects from its composite structure (game can be broken into separate components like rooms and obstacles).
We use Rainbow-CTS as a baseline: a Rainbow agent augmented with an intrinsic reward to handle the sparse rewards in \textsc{Montezuma's Revenge} \citep{bellemare2014skip}. Rainbow-CTS quickly achieves a game-score of 6600 but then plateaus even with additional training (Figure \ref{fig:f1}).

\begin{figure}[!t]
  \centering
  \subfigure[Performance ceiling for Rainbow-CTS agent.]
  {
  \includegraphics[width=0.25\columnwidth]{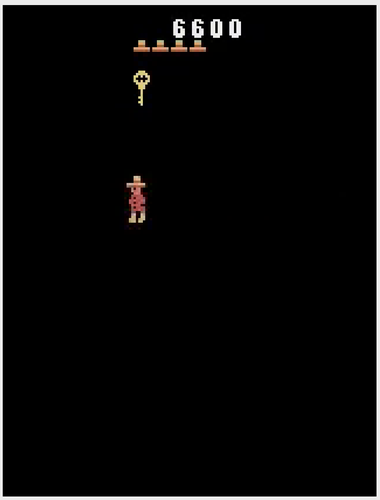}
  \label{fig: memento_6600}
  }
  \hfill
  \subfigure[Memento agent reliably makes further progress.]{\label{fig: memento_8000}\includegraphics[width=0.25\columnwidth]{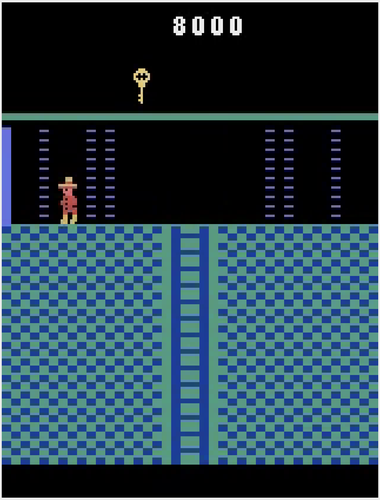}}
  \hfill
  \subfigure[Maximum score for Memento agent of 14500.]{\label{fig: memento_14500}\includegraphics[width=0.25\columnwidth]{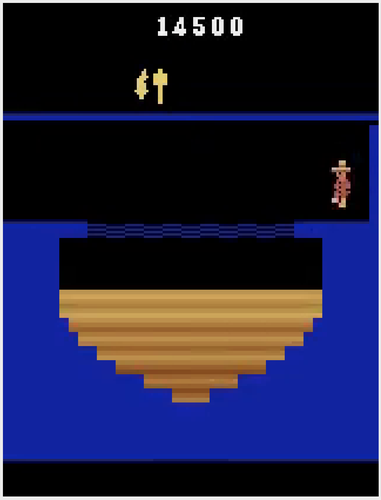}}
  \caption{\textbf{Comparing learning progress through the score:} A Rainbow-CTS agent fails to achieve game scores beyond 6660 (left) and remains stuck in this room with further training. However, an identical cloned Rainbow-CTS (Memento agent) launched from this position reliably makes further progress from here (middle) and repeated resets yielded maximum scores of 14500 (right).}
  \label{fig: mr_trajectory}
  \vspace{-.5em}
\end{figure}

We test the hypothesis that the Rainbow-CTS agent stagnates due to interference.
If the agent is affected by interference, when the agent models the game beyond the current plateau, this leads to interfering updates which negatively impacts the value function (and corresponding policy) earlier in the game. If this agent could learn about the frontier without interfering with the policy leading up to the frontier, we would therefore expect the agent to progress further. Therefore, we explicitly control for interference by launching a new, but identical agent (the Memento agent) from this position, and test whether the new agent can make further progress.

The Memento agent, launched from the state where the Rainbow-CTS agent plateaus, reliably escapes the plateau with a maximum score of 14,500 in \textsc{Montezuma's Revenge} (Figure \ref{fig:f2}, Figure \ref{fig: mr_trajectory}.
Since it begins each of its episodes from the \emph{final position} of the original agent, the Memento agent trains only on states beyond the plateau and performs parameter updates independently of the original agent.
This decouples the learning process between the states before and after the plateau, by design enforcing non-interference.
Appendix \ref{app: additional_observations} shows that the Memento agent with Rainbow-CTS similarly improves in other hard exploration games like \textsc{Gravitar}, \textsc{Venture}, and \textsc{Private Eye}.

The performance of the Memento agent is unexplained by longer training or by additional network capacity (Appendix \ref{appendix: rainbow_cts}) -- neither enables Rainbow-CTS to exceed the score plateau of 6600.
Exploration is also not the limiting factor; the original agent's replay buffer contains experience beyond the plateau. Rather, the agent is unable to \emph{integrate} this new information and learn a value function in the frontier region without degrading performance in the regions before.
These results imply that the Rainbow-CTS agent plateaus in performance not due to exploration, but rather catastrophic interference.

\section{Generalizing Across Agents and Games}\label{sec: generalizing}
Section \ref{sec: observation} supports the interference hypothesis in hard exploration games, a regime anticipated to be plagued by interference effects.
We now generalize the Memento experiment to measure interference prevalence across the ALE game suite and other learning algorithms.
To do so, we must address one experimental detail: how to discover a plateau state automatically.

\begin{wrapfigure}[18]{r}{0.5\textwidth}
    \centering
    \includegraphics[width=\linewidth]{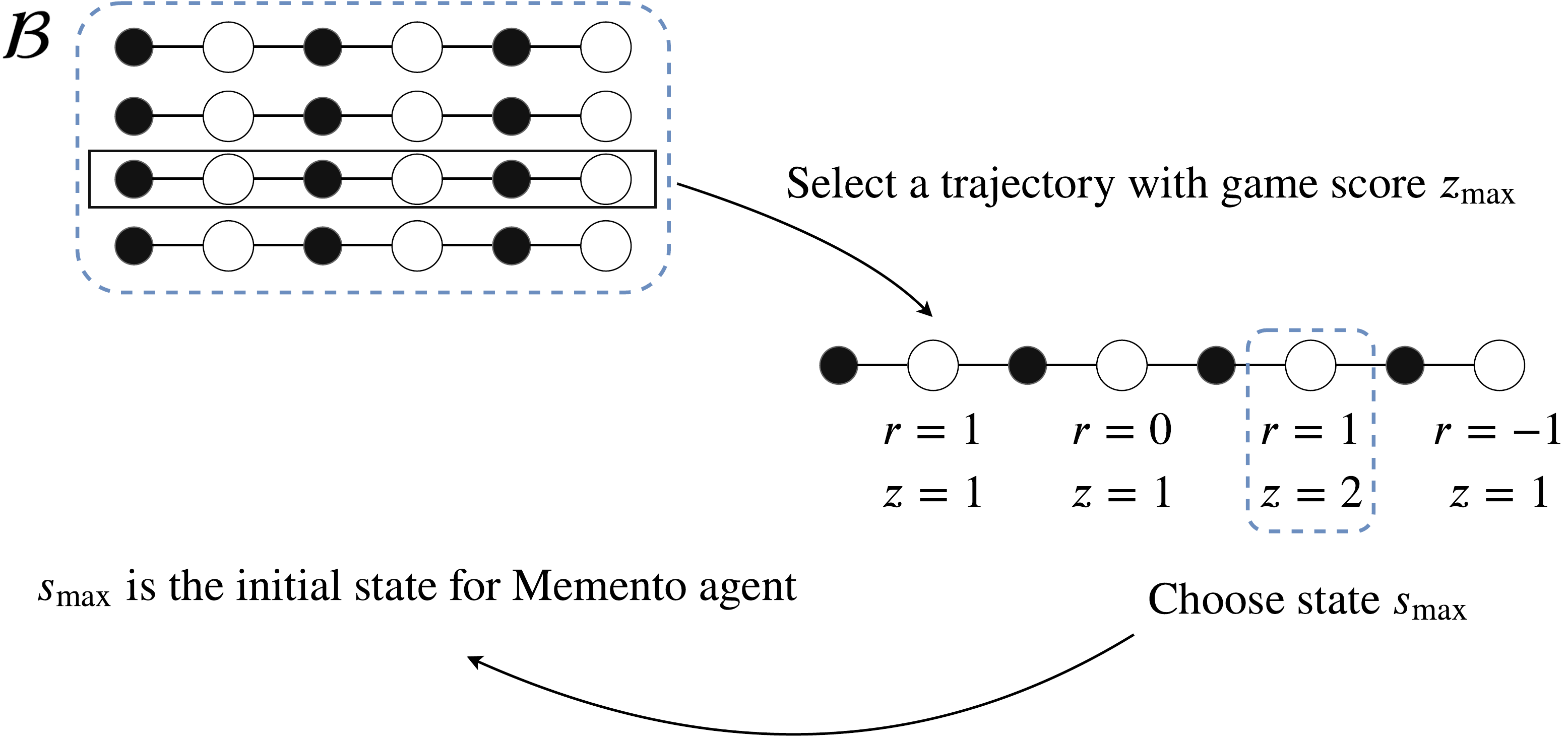}
    \caption{To find a plateau, we sample a trajectory from the replay buffer of the original trained agent. We compute the cumulative undiscounted return to date labeled $z$ for each state in the trajectory. We select the state $s_{\max}$ corresponding to the largest return to date ($z_{\max}$) to be the launch point for the Memento agent.}
    \label{fig: plateau_points}
\end{wrapfigure}
We adopt the definition that the \emph{plateau point} for an agent is the state where the agent achieves maximal game-score.
Figure \ref{fig: plateau_points} details our approach to do this. 
We first sample a trajectory from a trained agent's replay buffer, $(s_0, a_0, r_0, \cdots, s_{T}, a_{T}, r_{T})$.
We then compute the return-till-date (game score) for each state, $z_t = \sum_{k < t} r_k$, and choose the \emph{earliest} time-step $t_{\max}$ at which the game score is maximized: $t_{\max} = \min \{t : z_t = z_{\max}\}$.
The state at this time-step, $s_{\max} = s_{t_{\max}}$ will become the starting position for the Memento agent.
In Appendix \ref{appendix: many_states} we present alternative experiments launching the Memento agent from a diverse \emph{set} of states $\mathcal{S}_{\max}$ consistent with the maximum score and find that the results remain similar.

\begin{figure}[ht]
    \centering
    \includegraphics[width=\linewidth]{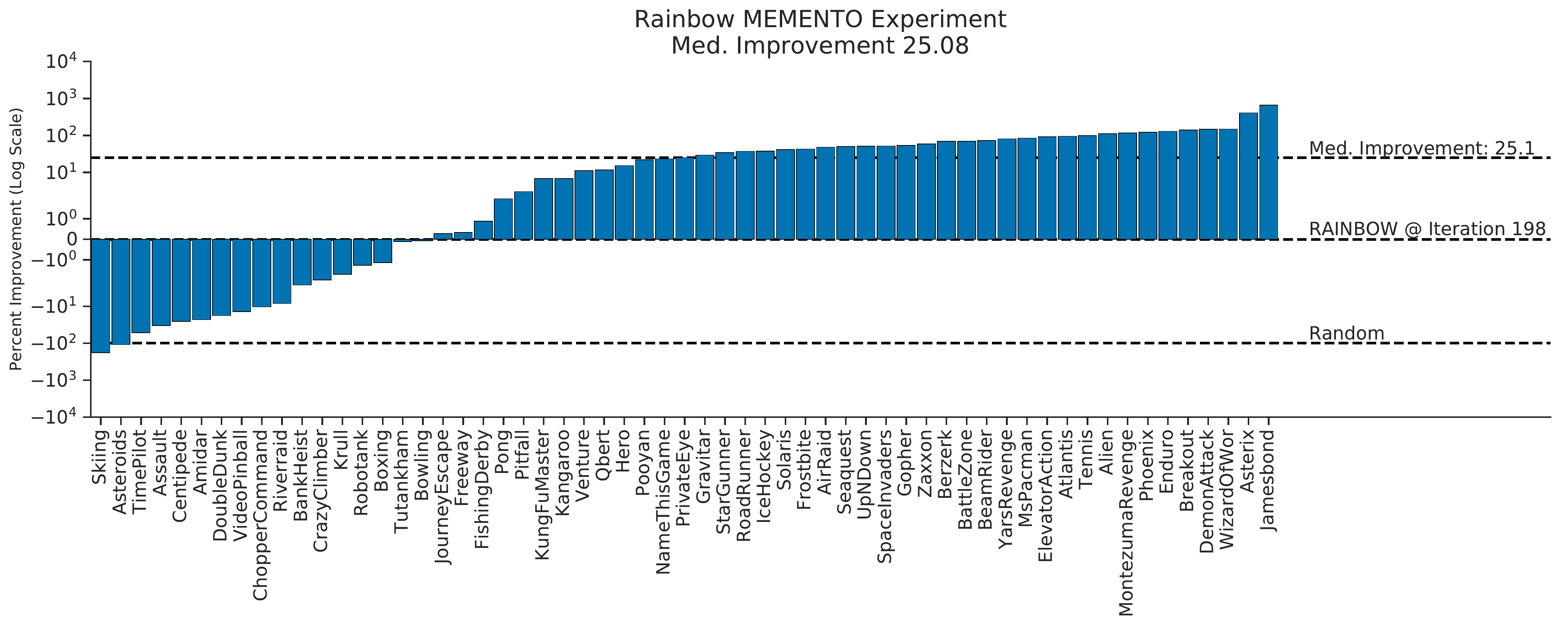}
    \caption{Each bar corresponds to a different game and the height is the percentage increase of the Rainbow Memento agent over the Rainbow baseline.  Across the entire ALE suite, the Rainbow Memento agent improves 75.0\% of the games performance with a median improvement of +25.0\%.}
    \label{fig: rainbow_memento}
\end{figure}

We conduct the Memento experiment with two different agents:  a Rainbow agent (without intrinsic motivation) and a DQN agent.
The Rainbow agent we use \citep{ castro2018dopamine} uses a prioritized replay buffer \citep{schaul2015prioritized}, n-step returns \citep{sutton1998reinforcement} and distributional learning \citep{bellemare2017distributional}.
Figure \ref{fig: rainbow_memento} shows that the Memento Rainbow agent achieves a significant +25.0\% median improvement over the original agent across the entire ALE suite.
This phenomenon also holds when ablating the three included algorithmic improvements of Rainbow and examining only the base DQN agent.
The DQN Memento agent achieves a $+17\%$ median improvement over the baseline agent (see Figure \ref{fig: dqn_memento} for game-by-game detail), supporting that these dynamics hold across different types of value-based agents.

We find that interference inhibits performance in a majority of Atari games, including \textsc{Asterix} and \textsc{Breakout} -- games without the compositional narrative structure of \textsc{Montezuma’s Revenge} and other hard exploration games.
This evidence supports the ubiquity of these interference issues.
In some cases, the Memento agent fails to improve, but this occurs mostly due to an poorly-chosen start state chosen by the heuristic. We did not attempt to refine this heuristic because the simple criterion had already produced sufficient evidence for the interference issues (though further performance could likely be achieved through more sophisticated state selection).

\begin{figure}%
    \centering
    \includegraphics[width=0.8\columnwidth]{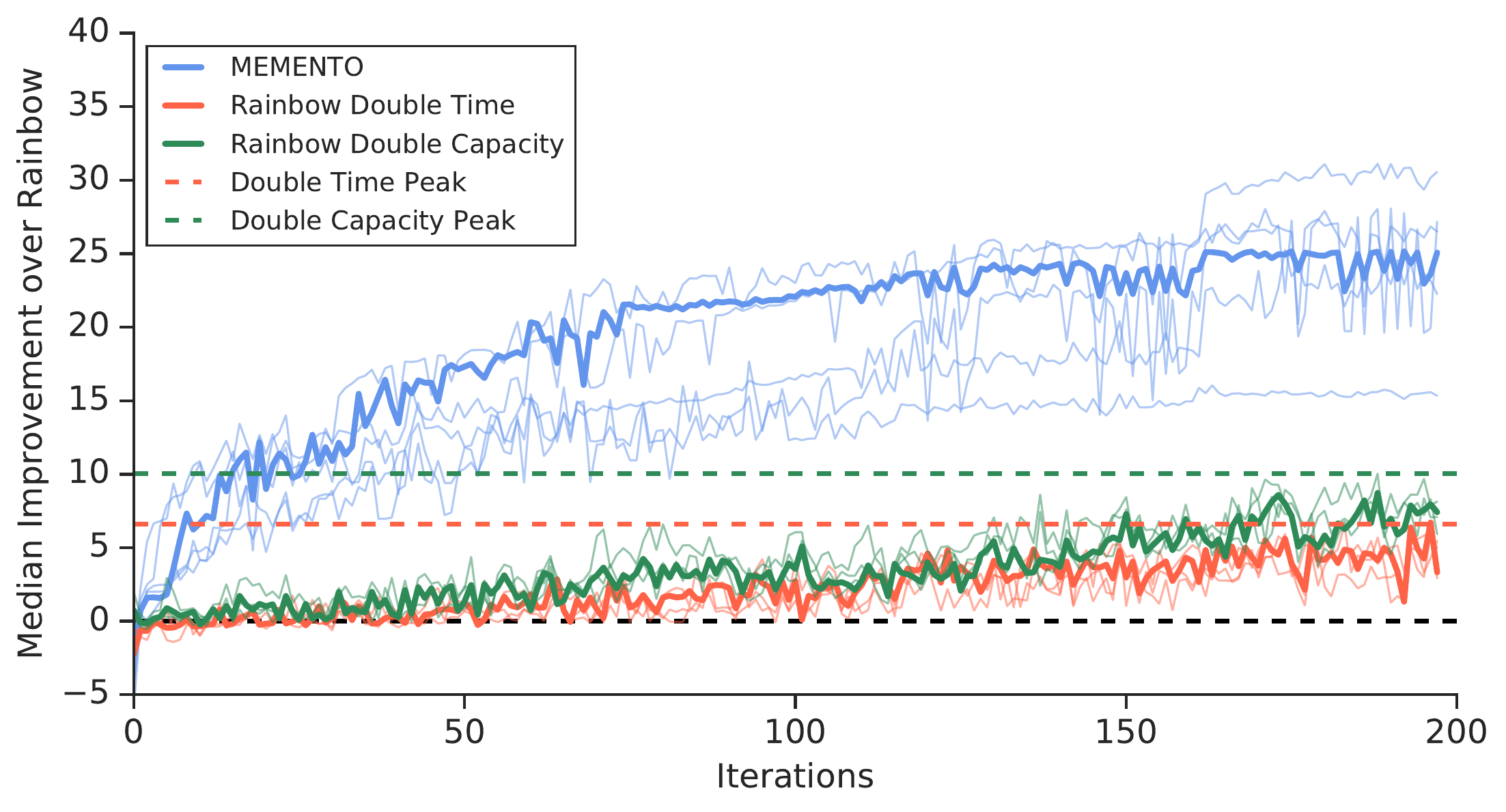}
    \caption{Performance of the Memento agent (blue) versus a standard Rainbow agent trained for 400M frames (red) and a  Rainbow agent with double network capacity trained for 400M frames (green).  We find that the Memento agent is considerably more sample efficient.  It reaches the peak performance of both longer-training Rainbow agents achieved over 200M frames in <25M frames.}
    \label{fig: memento_single_og_vs_trainlonger}
\end{figure}

\textbf{Discussion.}  As before, the longer training duration and additional model capacity of the Memento agent are insufficient to explain the performance boost.
We find in Figure \ref{fig: rainbow_duration} that training the original agent for 400M frames (double the conventional duration) results only in a +3.3\% median improvement for the Rainbow agent (versus +25.0\% in the Memento agent). 
Next in Figure \ref{fig: rainbow_capacity} we find that increased model capacity is an insufficient explanation of the Memento agent's performance by training a double capacity Rainbow agent and finding an improvement of only +7.4\% (versus +25.0\% for the Memento agent). 
Figure \ref{fig: memento_single_og_vs_trainlonger} shows how the performance gap widens over training duration.
Furthermore, the Memento agent is far more \emph{sample efficient} than a baseline Rainbow agent that is trained for longer.
This new agent achieves the same median performance increase in only 5M frames compared to 200M frames  for the longer-training Rainbow agent.

\section{Measuring Intra-Game Catastrophic Interference}
The prior section provided support for the hypothesis by demonstrating that two decoupled -- and thus non-interfering -- agents make further progress than a single agent afforded longer training duration or increased model capacity.
We now investigate this hypothesis through another lens and a finer level of detail: specifically, we measure the TD-errors in different segments of the game as the agent learns about other segments.  We center our analysis on two well-known Atari games where the Memento agent did not progress significantly (\textsc{Pong} and \textsc{QBert}) and two games where the Memento agent did improve (\textsc{Montezuma's Revenge} and \textsc{Breakout}).

Formally, we split a game into \emph{contexts} by using the game score as a task contextualization \citep{jain2019context}: each context corresponds to a segment of the game with the same game score.
While changes in game score are not a perfect demarcation of true boundaries in a game, they are an easily extracted proxy, and often represent significant milestones in a game. We now propose an approach to measure catastrophic interference between contexts in a game.

During training, a value network with parameters $\theta$ is trained to minimize the temporal difference (TD) error $\delta_\theta(s,a,r,s') = r + \gamma \max_{a'\in\Acal} Q_{\tau}(s', a') - Q_{\theta}(s,a)$.
The TD error is averaged over transition samples $(s, a, s', a') \sim \mu(\mathcal{B})$ to form the non-distributional loss 
\begin{align*}
    \Lcal(\theta) = \Ebf_{\mu} \left[L\left(\delta_\theta(S,A,R,S')\right)\right],
\end{align*}
where $L$ refers to the Huber loss.
We refer readers to \citet{bellemare2017distributional} for the distributional loss.
In the interim period, while the target network is held fixed, the optimization of the online network parameters $\nabla_\theta \Lcal(\theta)$ reduces to a strictly \emph{supervised learning} setting, eliminating several complications present in RL.
The primary complication avoided is that for any given transition $(s, a, r, s')$ the Bellman targets do not change as the online network evolves.
In the general RL setting with bootstrapping through TD-learning, the targets will necessarily change, otherwise the online network will simply regress to the $Q$-values established with the initial target parameters.
We analyze this simplified regime in order to carefully examine the presence of catastrophic interference.

If the learning generalizes, then we predict TD-errors elsewhere to decrease.
However, if the learning interferes, then we predict TD-errors elsewhere to \emph{increase}.
In Appendix \ref{app: experience_replay} we provide intuition and evidence how experience replay can produce sampling dynamics that lead to catastrophic interference.

\begin{figure}%
    \centering
    \includegraphics[width=0.9\linewidth]{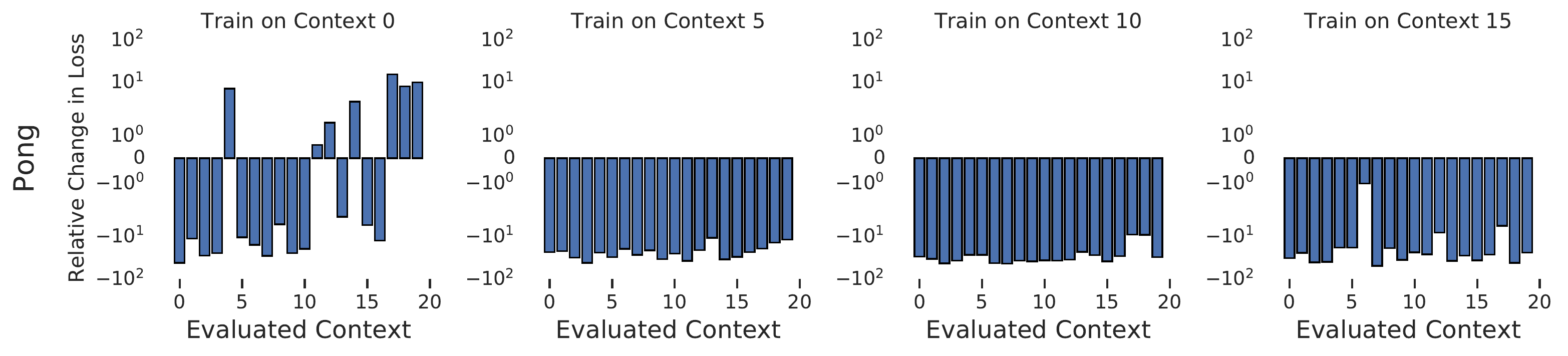}
    \includegraphics[width=0.9\linewidth]{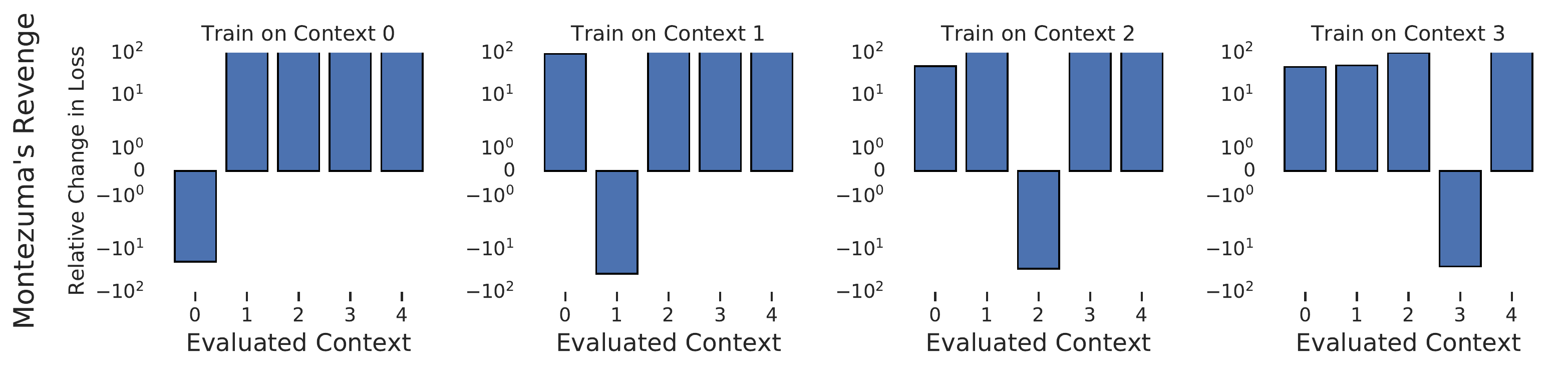}
    \caption{We track the relative changes in TD errors for \textit{all} contexts when a Rainbow agent is trained on a particular context on \textsc{Pong} and \textsc{Montezuma's Revenge}. In \textsc{Pong}, training on a context generally reduces the loss on all contexts uniformly. In contrast, on \textsc{Montezuma's Revenge}, training on any context leads to negative generalization and increased loss on all other contexts.}
    \label{fig: pong_td_error_analysis}
\end{figure}

\subsection{Low Catastrophic Interference.}
\begin{figure}%
    \centering
    \subfigure[\textsc{Pong}]{\label{fig: pong_colorbar}\includegraphics[width=0.33\linewidth]{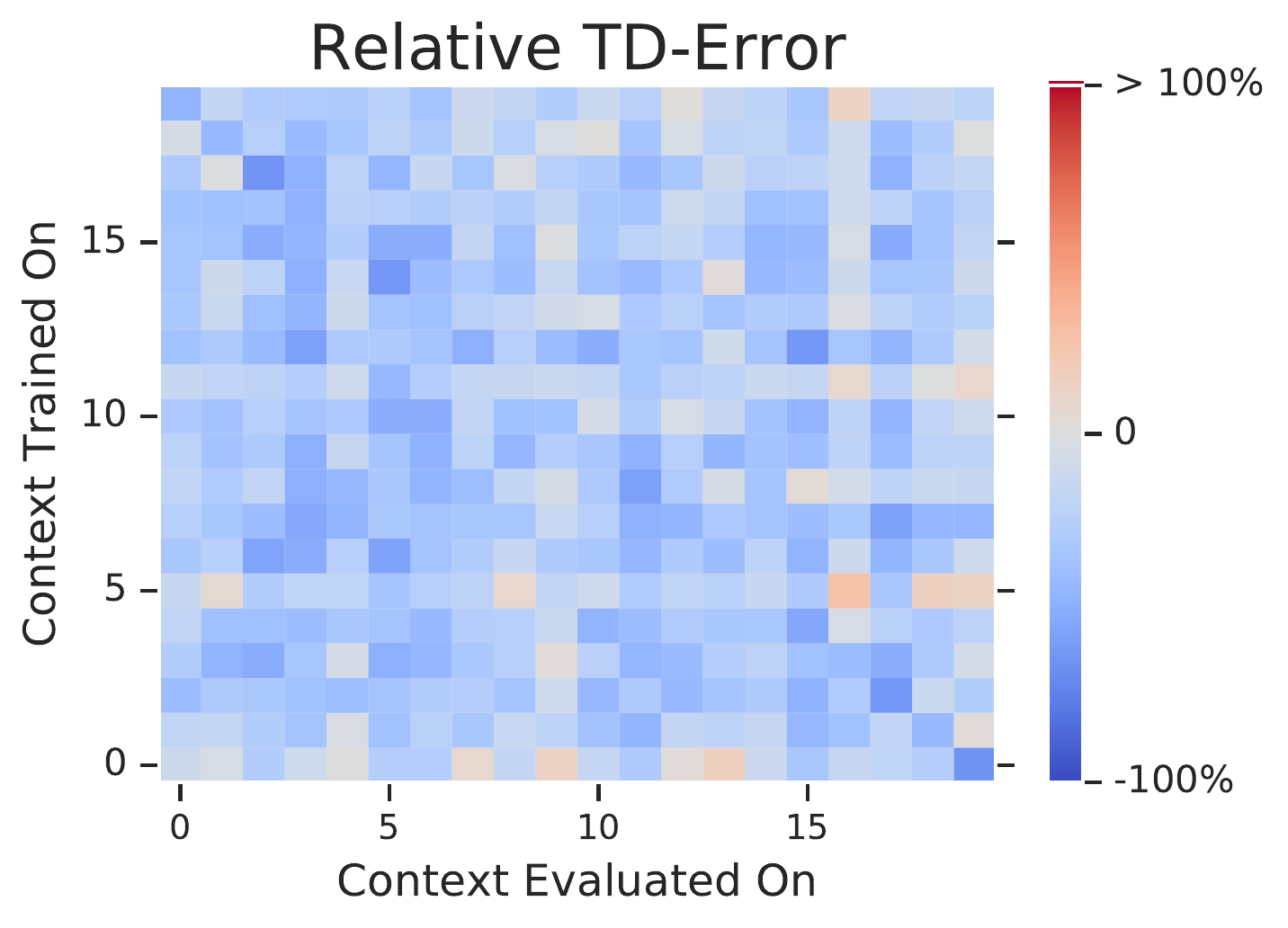}}
    \subfigure[\textsc{Montezuma's Revenge}]{\label{fig: colorbar}\includegraphics[width=0.36\linewidth]{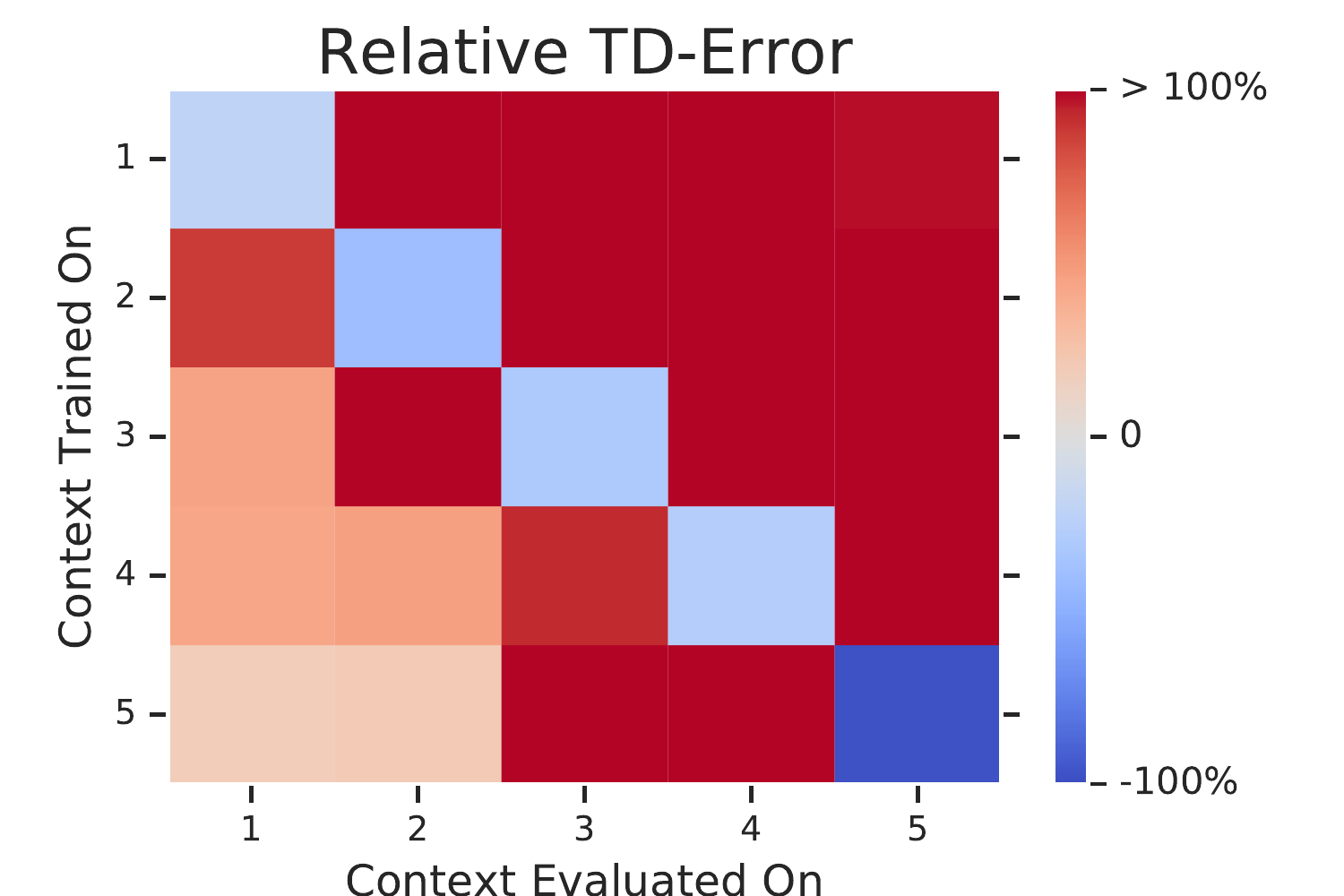}}
    \vfill
    \subfigure[\textsc{QBert}.]{\label{fig: qbert_colorbar}\includegraphics[width=0.36\linewidth]{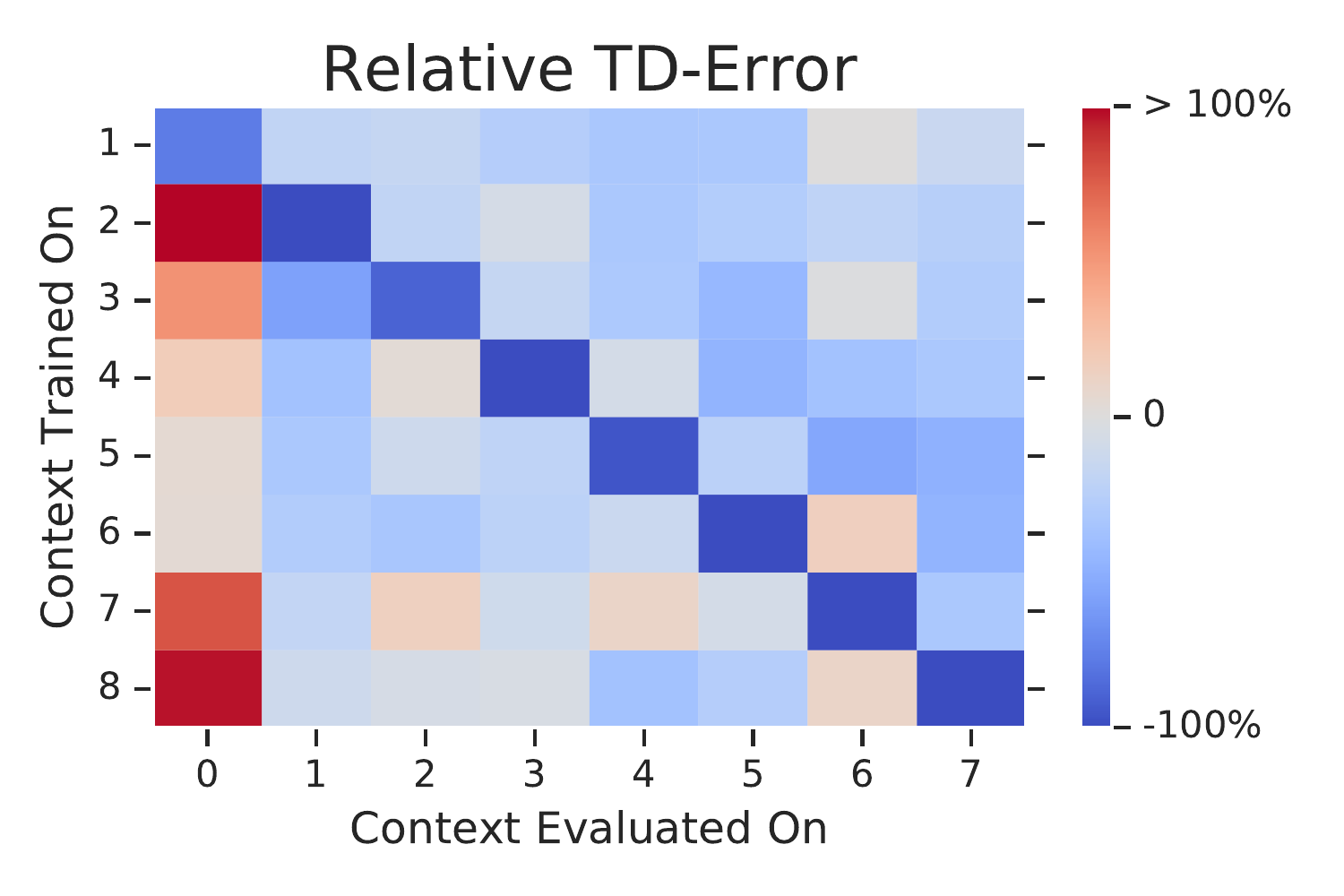}}
    \subfigure[\textsc{Breakout}.]{\label{fig: breakout_colorbar}\includegraphics[width=0.36\linewidth]{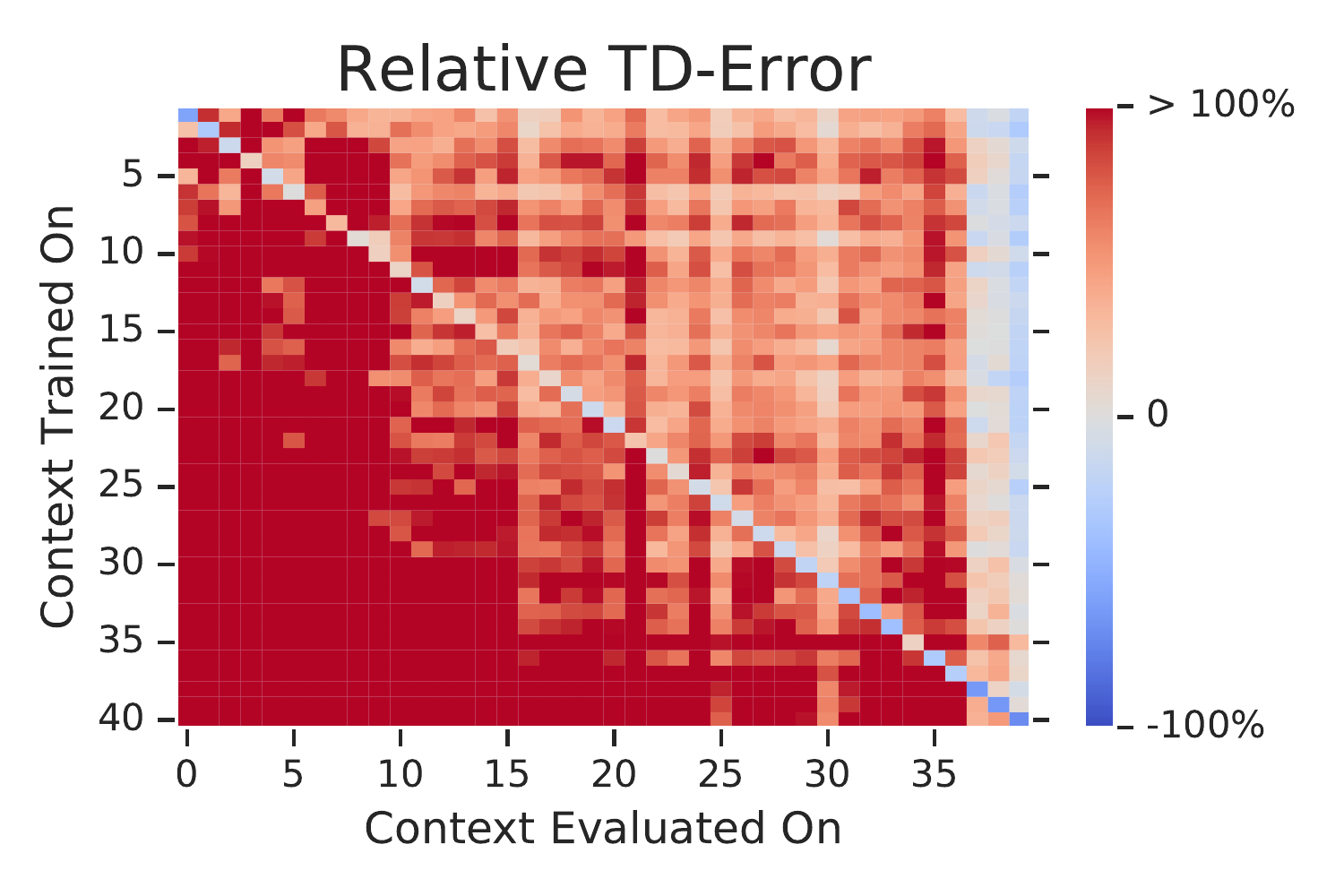}}
    \caption{Training on a particular context unpredictably changes TD errors elsewhere. Blue grids indicates generalization and red indicates interference between contexts. For \textsc{Pong} and \textsc{Qbert}, learning generalizes across contexts. However, for \textsc{Montezuma's Revenge} and \textsc{Breakout}, training leads to negative transfer between contexts with strong evidence of catastrophic interference.} 
\end{figure}

We first consider \textsc{Pong}, a game where the Memento agent does not improve for either Rainbow or DQN.
Figure \ref{fig: pong_td_error_analysis} visualizes relative changes in TD errors across all contexts when trained on other contexts.
Learning in one context or game score generalizes very well to others (with occasional aberration) -- training on any context generally reduces TD errors elsewhere.
Figure \ref{fig: pong_colorbar} provides a compressed view of the relative changes in TD-errors, where the $(i,j)$-th entry corresponds to the relative change in TD error for context $j$, when the model is trained on context $i$. Locations where the diagram is \emph{red} correspond to interference and negative generalization, and \emph{blue} regions correspond to positive generalization between contexts. The diagram is overwhelmingly blue, indicating positive generalization between most regions. Similarly, in \textsc{Qbert}, another game where the Memento agent did not progress significantly, training on a specific context typically generalizes to other contexts (Figure \ref{fig: qbert_colorbar}). However, there is consistent negative impact for the 0th context, suggesting that the beginning of the game (context 0) differs from later sections. 

\subsection{High Catastrophic Interference.}
We have observed learning about one portion of \textsc{Pong} or \textsc{Qbert} typically improves the prediction error in other contexts -- \emph{learning generalizes}. However, this is not generally the case.
In \textsc{Montezuma's Revenge}, where the Memento agent progresses significantly, we examine interference amongst the first five game score contexts, corresponding to milestones including collecting the first key and exiting the initial room. In contrast to \textsc{Pong}, learning about any context in \textsc{Montezuma's Revenge} strictly \emph{interferes} with the prediction errors in other contexts (Figure \ref{fig: pong_td_error_analysis}).

Figure \ref{fig: colorbar} shows the corresponding TD-error reduction matrix; as expected, training on samples from a particular context reduces the TD-errors of that context as observed by the blue diagonals.
However, the red off-diagonals indicate that training on samples from one context negatively \emph{interferes} with predictions in all the other contexts.
Issues of catastrophic forgetting also extend to games where it might be unexpected like \textsc{Breakout}.
Figure \ref{fig: breakout_colorbar} shows that a Rainbow agent trained here interferes unpredictably with other contexts. In \textsc{Breakout}, we note an interesting asymmetry. Training on later contexts produces worse backward transfer (bottom-left) while training on earlier contexts does not produce as severe forward transfer (top-right) \citep{lopez2017gradient}.

\section{Related Work}
\paragraph{Catastrophic Interference:} To control how memories are retained or lost, many study the synaptic dynamics between different environment contexts. These methods either try to regularize or expand the parameter space. Regularization has shown to discourage weight updates from overwriting old memories. Elastic Weight Consolidation (EWC) \citep{kirkpatrick2017overcoming} employs a quadratic penalty to keep parameters used in successive environments close together, and thus prevents unrelated parameters from being repurposed. \citet{li2017learning} impose a knowledge distillation penalty \citep{hinton2015distilling} to encourage certain network predictions to remain similar as the parameters evolve. The importance of some memories can be encoded in additional weights, then used to scale penalties incurred when they change \citep{zenke2017continual}. In general, regularization is effective when a low loss region exists at the intersection of each context’s parameter space. When these regions are disjoint, however, it can be more effective to freeze weights \citep{sharif2014cnn} or to replicate the entire network and augment it with new features for new settings \citep{rusu2016progressive, yoon2018lifelong, draelos2017NeurogenesisDL}. Our work studies model expansion in the context of Atari, with a hypothesis that little overlap exists between the low loss regions of different game contexts. However, some games, such as Pong, exhibit considerable overlap. 

Knowledge of the environment context is often a prerequisite to regularization or model expansion. Given a fixed strategy to mitigate interference, some consider the problem finding the best contextualization. \citet{rao2019continual} proposed a variational method to learn context representations which apply to a set of shared parameters. Similar to the Forget-Me Not Process \citep{milan2016fmn}, the model is dynamically expanded as new data is experienced which cannot be explained by the model. \citet{aljundi2019taskfreecl} proposes a similar approach using regularization.
Our work posits a context model that marks the boundaries between environment settings with the game score.

\paragraph{Continual Learning:}
Multi-task and continual learning in reinforcement learning is a highly active area of study \citep{ring1994continual, silver2013lifelong, oh2017zero} with some proposing modular architectures, however, multi-task learning in the context of a single environment is a newer area of research.
\citet{schaul2019ray} describe the problem of ray interference in multi-component objectives for bandits, and show that in this setting, learning exhibits winner-take-all dynamics with performance plateaus and learning constrained to only one component.
In the context of diverse initial state distributions, \citet{ghosh2017divide} find that policy-gradient methods myopically consider only a subset of the initial-state distribution to improve on when using a single policy.
The Memento observation makes close connections to Go-Explore  \citep{ecoffet2019go} which demonstrates the efficacy of resetting to the boundary of the agent's knowledge and thereafter employing a basic exploration strategy to make progress in difficult exploration games.
Finally, recent work in unsupervised continual learning consider a case similar to our setting, where the learning process is effectively multitask but there are no explicit task labels \citep{rao2019continual}.

\section{Discussion}
This research provides evidence of a difficult \emph{continual learning problem} arising within a single game.
The challenge posed by catastrophic interference in the reinforcement learning setting is greater than in standard continual learning for several reasons.
First, in this setting we are provided no explicit task labels \citep{rao2019continual}.
Our analyses used as a proxy label, the game score, to reveal continual learning dynamics, but alternative task designations may further clarify the nature of these issues.
Next, adding a further wrinkle to the problem, the "tasks" exist as part of a single MDP and therefore generally \emph{must} transmit information to other tasks.
TD errors in one portion of the environment directly effect TD-errors elsewhere via Bellman backups \citep{bellman1957markovian, sutton1998reinforcement}.
Finally, although experience replay \citep{mnih2013playing, mnih2015human} resembles shuffling task examples, we find it is not sufficient to combat catastrophic forgetting unlike the standard setting. We posit that this is likely due to complicated feedback effects of data generation and learning.

\section{Conclusion}
We establish empirical connections between catastrophic forgetting and central issues in reinforcement learning including poor sample efficiency, performance plateaus \citep{schaul2019ray} and exploration challenges \citep{taiga2019benchmarking}.
\citet{schaul2019ray} hypothesized interference patterns might be observed in a deep RL setting for games like \textsc{Montezuma's Revenge}.
Our empirical studies confirm this hypothesis,and further show that the phenomenon is more prevalent than previously conjectured.
Both the Memento experiment as well as peering into inter-context interference illuminates the nature and severity of interference in deep reinforcement learning.
Our findings also suggest that the prior belief of what constitutes a "task" may be misleading and must therefore be carefully examined. 
We hope this work provides a clear characterization of the problem and show that it has far reaching implications for many fundamental problems in reinforcement learning.

\section*{Acknowledgements}
This work stemmed from surprising initial results and our understanding was honed through many insightful conversations at Google and Mila.
In particular, the authors would like to thank Marlos Machado, Rishabh Agarwal, Adrien Ali Ta{\"\i}ga, Margaret Li, Ryan Sepassi, George Tucker and Mike Mozer for helpful discussions and contributions.
We would also like thank the reviewers at the Biological and Artificial Reinforcement Learning workshop for constructive reviews on an earlier manuscript.

\bibliography{references}
\bibliographystyle{icml2020}

\clearpage
\appendix
\section{Experience Replay}\label{app: experience_replay}
\begin{figure}[ht]
    \centering
    \includegraphics[width=0.6\linewidth]{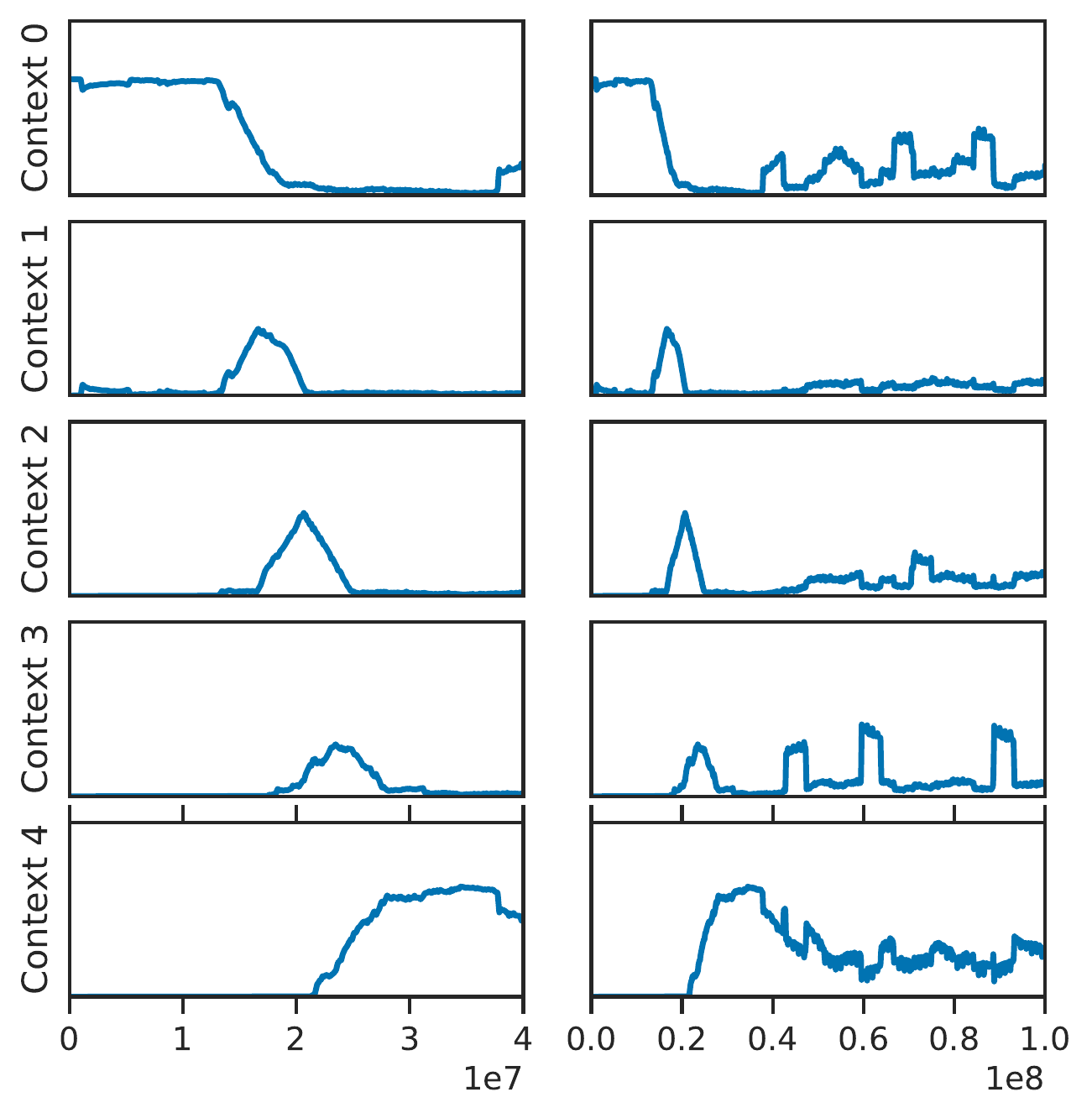}
    \caption{We plot how often the first five game contexts of \textsc{Montezuma's Revenge} are sampled from the replay buffer throughout training. Early in training (left column) the agent almost exclusively trains on the first stage (because no later stages have been discovered). In intermediary stages, the agent is being trained on all the contexts at the same time (which can lead to interference), and in late stages, is being trained on only the last stage (which can lead to catastrophic forgetting).  After discovering all context (right column), the agent oscillates between sampling contexts.}
    \label{fig: multi_task_rl}
\end{figure}

Replay buffers with prioritized experience replay \citep{schaul2015prioritized}, which preferentially samples states with higher temporal-difference (TD) error, can exasperate catastrophic interference.
Figure \ref{fig: multi_task_rl} records the training dynamics the Rainbow agent in \textsc{Montezuma's Revenge}.
Each plot is a separate time-series recording which context(s) the agent is learning as it trains in the environment.
The agent clearly iterates through stages of the environment indexed by the score, starting by learning exclusively context 0, then context 1 and so on.
The left column shows the early learning dynamics and the right plot shows resulting oscillations after longer training.
This is antithetical to approaches to address catastrophic forgetting that suggest \emph{shuffling} across tasks and contexts \citep{ratcliff1990connectionist, robins1995catastrophic}. 

These dynamics are reminiscent of continual learning\footnote{In contrast to the standard setting, these score contexts can't be learned \emph{independently} since the agent must pass through earlier contexts to arrive at later contexts.} where updates from different sections may interfere with each-other \citep{schaul2019ray}.
Prioritized experience replay, a useful algorithm for sifting through past experience, and an important element of the Rainbow agent, naturally gives rise to continual learning dynamics.
However, removing prioritized experience replay and sampling instead randomly and uniformly \citep{mnih2013playing, mnih2015human} can still manifest as distinct data distribution shifts in the replay buffer.
As the agent learns different stages of the game, frontier regions may produce more experience as the agent dithers in exploratory processes.

\section{Rainbow Agent with CTS Additional Figures}
\subsection{Additional Training and Model Capacity}\label{appendix: rainbow_cts}
We consider training the Rainbow CTS model longer and with higher capacity.

\begin{figure}[!h]
    \centering
    \subfigure[Result of training the Rainbow CTS for a longer duration. The black vertical line denotes the typical training duration of 200M frames.  We find for both the original agent (orange) as well as a sweep over other update-horizons (blue, green), that no agent reliably exceeds the baseline.]{\label{fig: long_duration}\includegraphics[width=0.4\columnwidth]{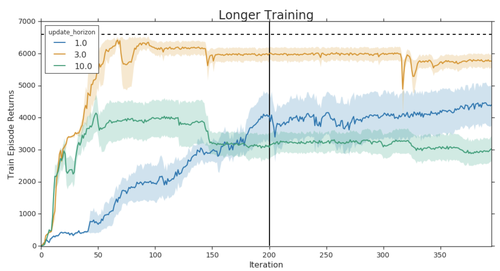}}
    \hfill
    \subfigure[Result of training a Rainbow CTS with the increased capacity equal to two separate Rainbow CTS agents accomplished by increasing the filter widths.  We find for both the original agent (orange) as well as a sweep over other update-horizons (blue, green), that no agent reliably exceeds the baseline.]{\label{fig: more_capacity}\includegraphics[width=0.4\columnwidth]{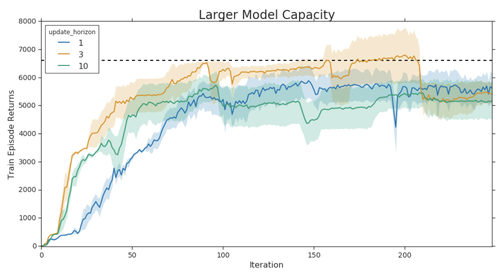}}
     \caption{In \textsc{Montezuma's Revenge}, neither additional training (Figure \ref{fig: long_duration}) nor additional model capacity (Figure \ref{fig: more_capacity}) leads to improved performance from the base agent. These results hold for various n-step returns or update horizons of 1, 3 (Rainbow default), and 10. The dotted-line is the maximum achieved baseline for the original agent, and all experiments are run with 5 seeds.}
    \label{fig: memento_duration}
\end{figure}

\subsection{Rainbow with CTS Hard Exploration Games}
\label{app: additional_observations}
We demonstrate the Memento observation for three difficult exploration games using the Rainbow + CTS agent.

\begin{figure}[!h]
    \centering
    \subfigure[The Memento observation in \textsc{Gravitar}.]{\includegraphics[width=0.3\columnwidth]{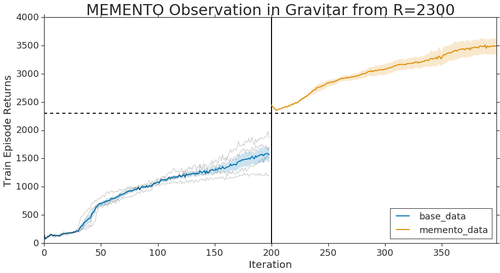}}
    \hfill
    \subfigure[The Memento observation in \textsc{Venture}.]{\includegraphics[width=0.3\columnwidth]{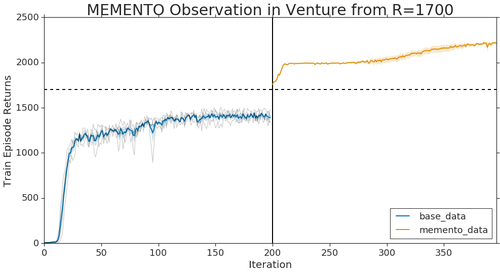}}
    \hfill
    \subfigure[The Memento observation in \textsc{PrivateEye}.]{\includegraphics[width=0.3\columnwidth]{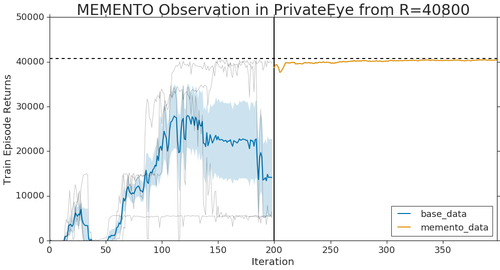}}
    \caption{Each black line represents a run from each of the five seeds and the dotted black line is the maximum achieved score.  We find that the Memento observation holds across hard exploration games.  \textsc{PrivateEye} results in a slight decrease in performance for the new agent (orange) due to the count-down timer in, however, we note that it is more stable compared to baseline data (blue) and preserves performance.}
\end{figure}

\clearpage
\section{Rainbow Agent Additional Figures}

\subsection{Additional Training and Model Capacity}
We first present results of training the Rainbow agent for a longer duration and with additional capacity.
In both settings, we find these variants fail to achieve the performance of the Memento observation.

\begin{figure}[!ht]
    \centering
    \subfigure[A Rainbow agent trained for twice the duration or 400M frames.]{\label{fig: rainbow_duration}\includegraphics[height=12cm]{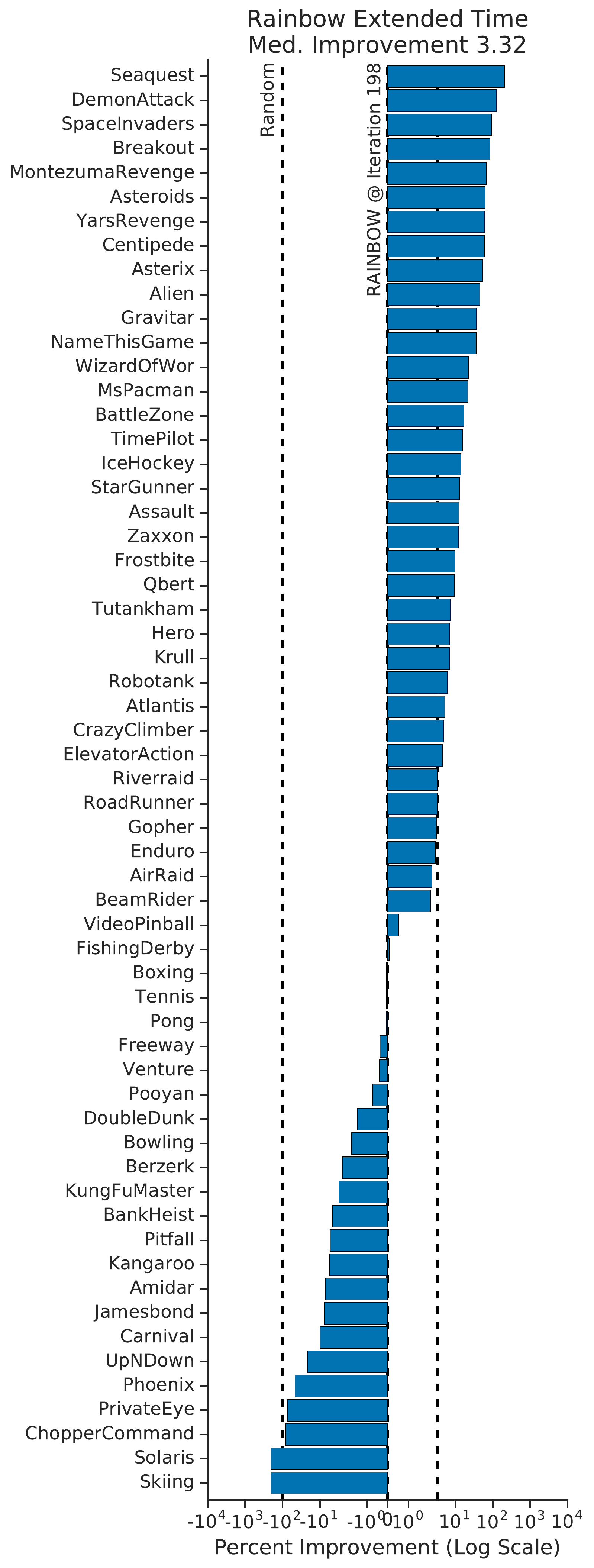}}
    \subfigure[A Rainbow agent trained with double network capacity.]{\label{fig: rainbow_capacity} \includegraphics[height=12cm]{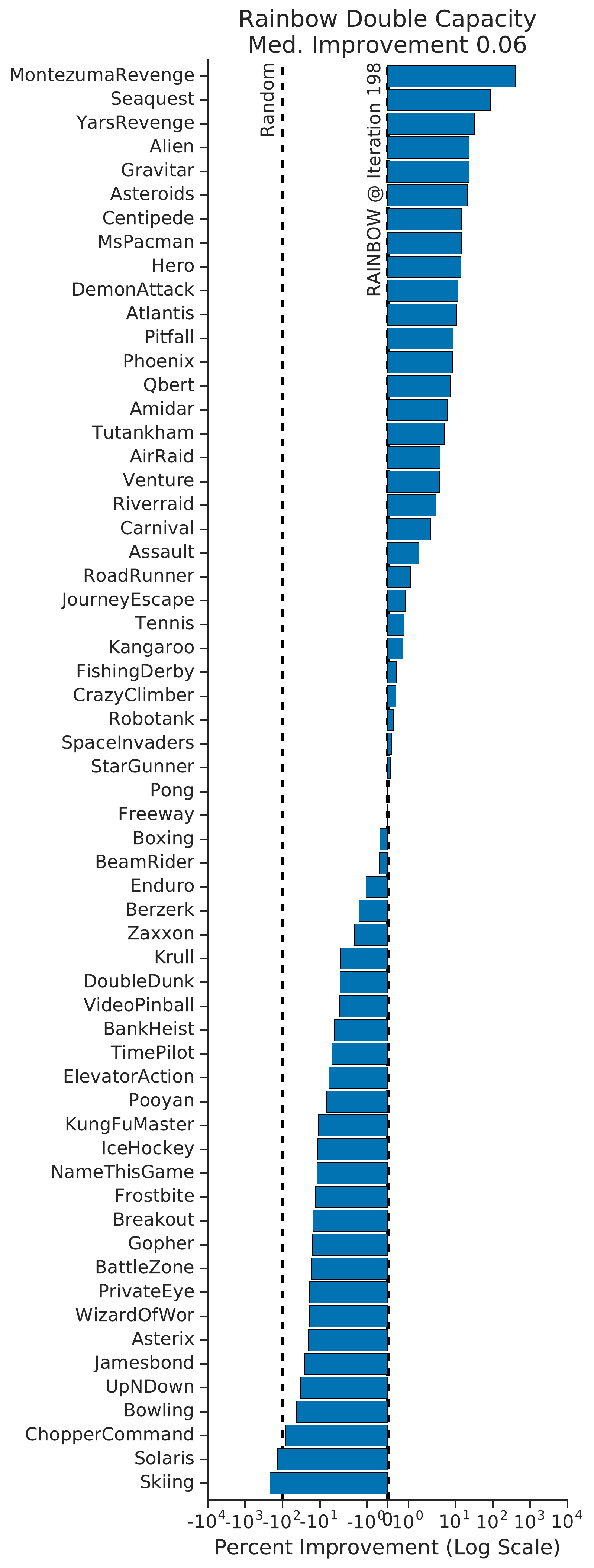}}
    \caption{A Rainbow agent trained for twice the duration or twice the network capacity similarly fails to improve materially over the baseline.  Training for 400M frames yields a median improvement of +3\% and doubling the network capacity yields no median improvement.  In the contrast, the Memento agent improves over baseline +25.0\%.}
\end{figure}

\clearpage
\section{Memento State Strategy}\label{appendix: many_states}
We consider the performance when the Memento agent instead starts from a \emph{set} of states $\mathcal{S}$ rather than a singular state.
This is a more difficult problem requiring demanding more generalization capability from the Memento agent and thus reported median performance is lower.
We find, however, that the Memento observation also holds in this more challenging environment.
Furthermore, we find a higher fraction of games improve under multiple Memento start states, likely due to avoiding the primary issue of starting exclusively from an inescapable position.

\begin{figure}[!h]
    \centering
    \subfigure[Single Memento state.]{\includegraphics[height=12cm]{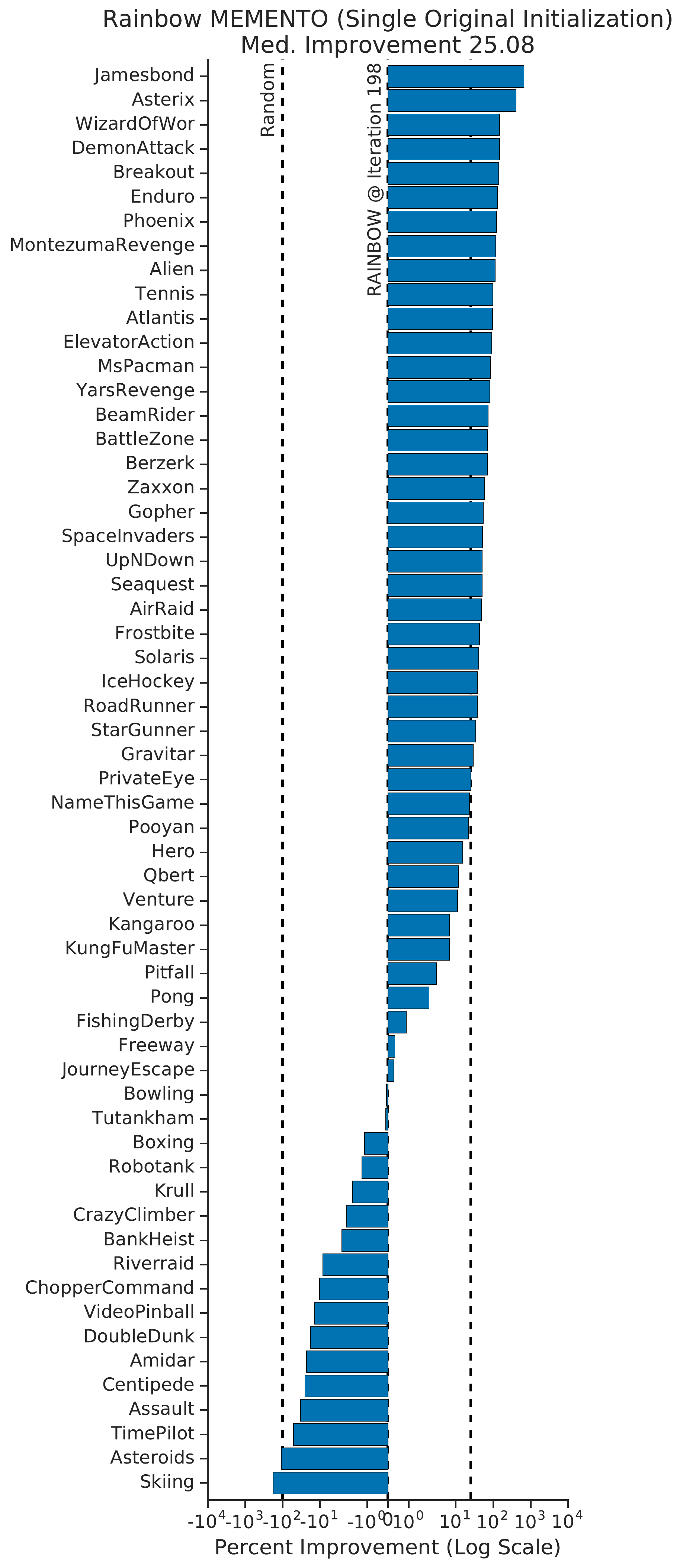}}
    \subfigure[Multiple Memento states.]{\includegraphics[height=12cm]{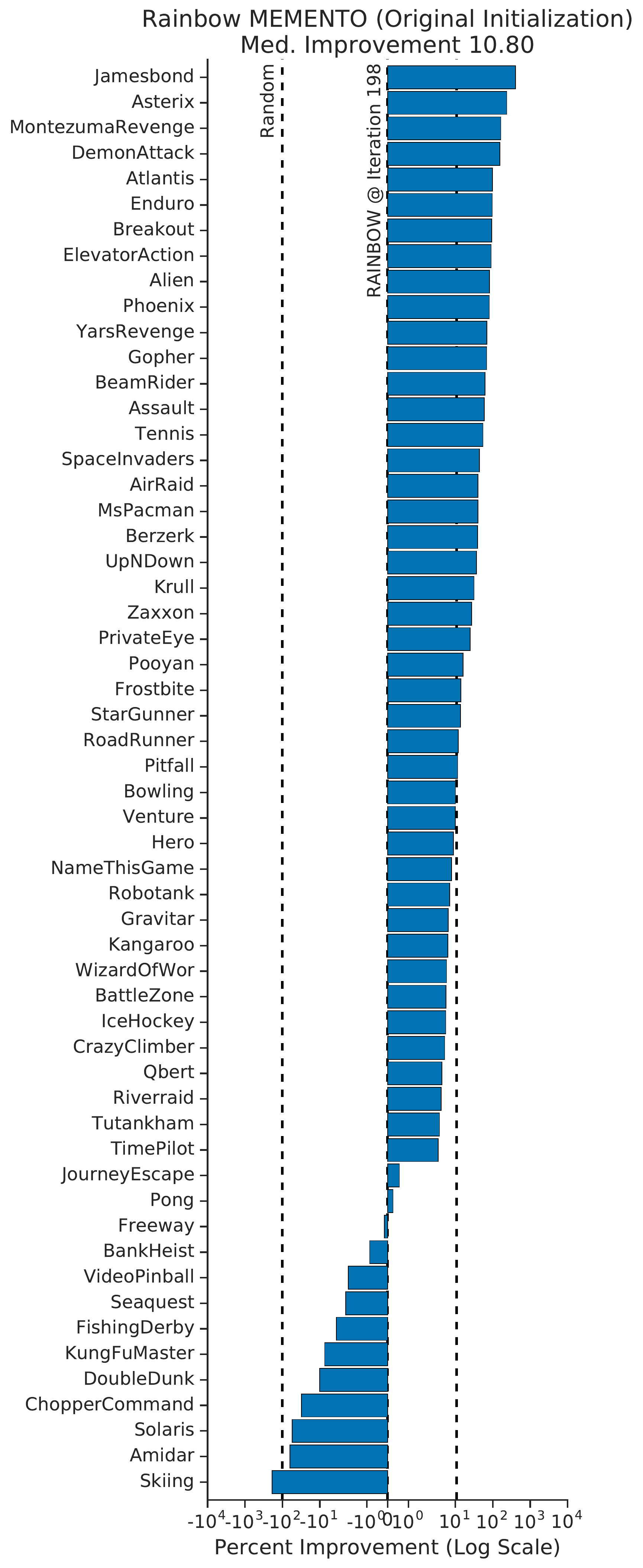}}
    \caption{\textbf{Multiple Memento states.} Each bar corresponds to a different game and the height is the percentage increase of the Rainbow Memento agent from a set of start states over the Rainbow baseline.  The Rainbow Memento agent launched from a \emph{single} state results in a 25.0\% median improvement (left) which is greater than the 11.9\% median improvement of launching from \emph{multiple} states (right).}
\end{figure}

\clearpage
\section{Weight Transfer}
We examine the importance of the original agent's parameters in order transfer to later sections of the game in the Memento observation.
Our original observation relied on the \emph{original agent} weights.
To examine the weight transfer benefits of these parameters, we consider instead using \emph{random initialization} for the Memento agent.
If the parameters learned in the early section of the game by the original agent generalize, we should observe faster progress and potentially better asymptotic performance.
We measure this both for Rainbow and DQN.

\begin{figure}[!h]
    \centering
    \subfigure[Rainbow Memento agent.]{\includegraphics[width=0.45\columnwidth]{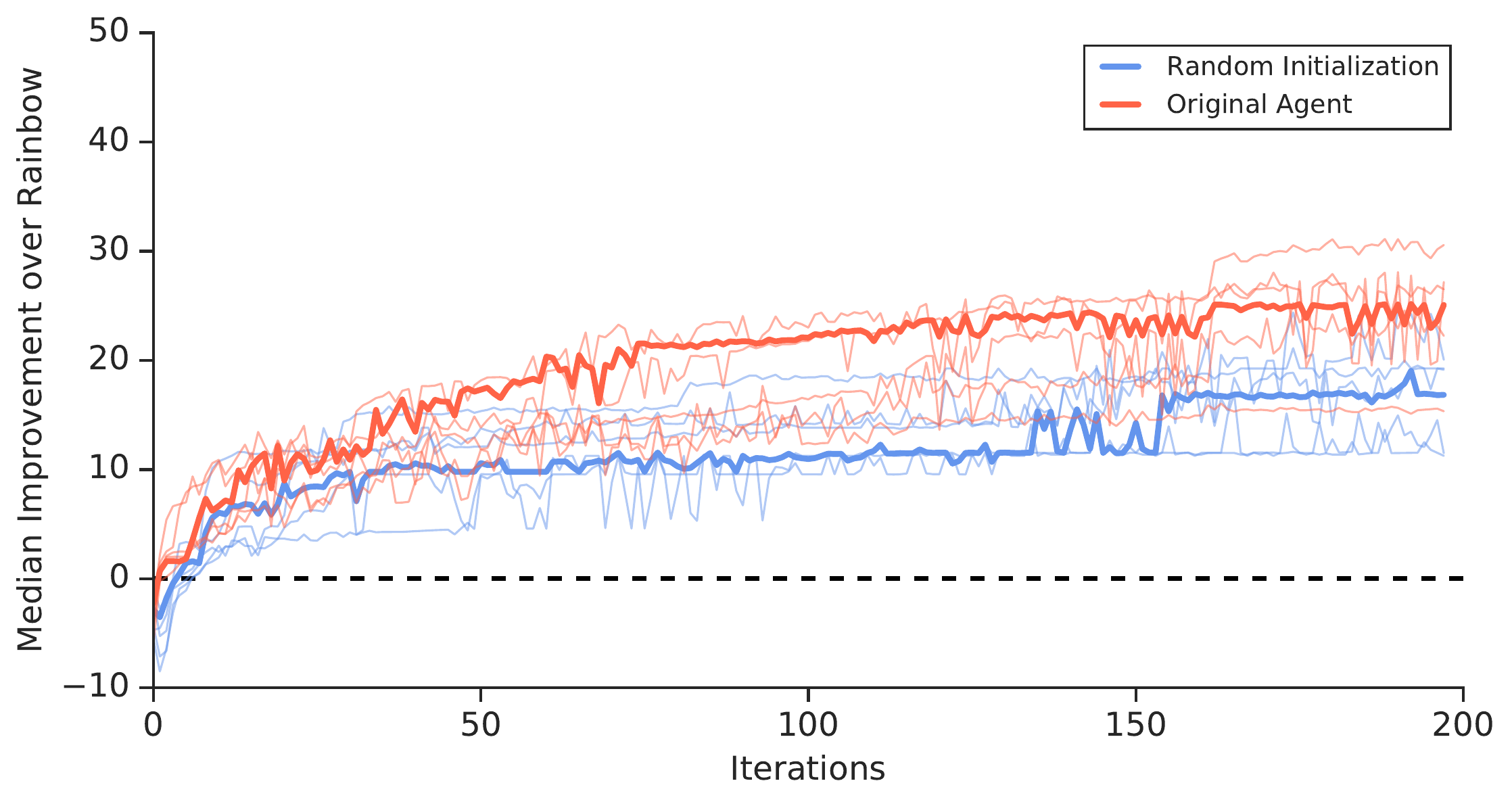}}
    \subfigure[DQN Memento agent.]{\includegraphics[width=0.45\columnwidth]{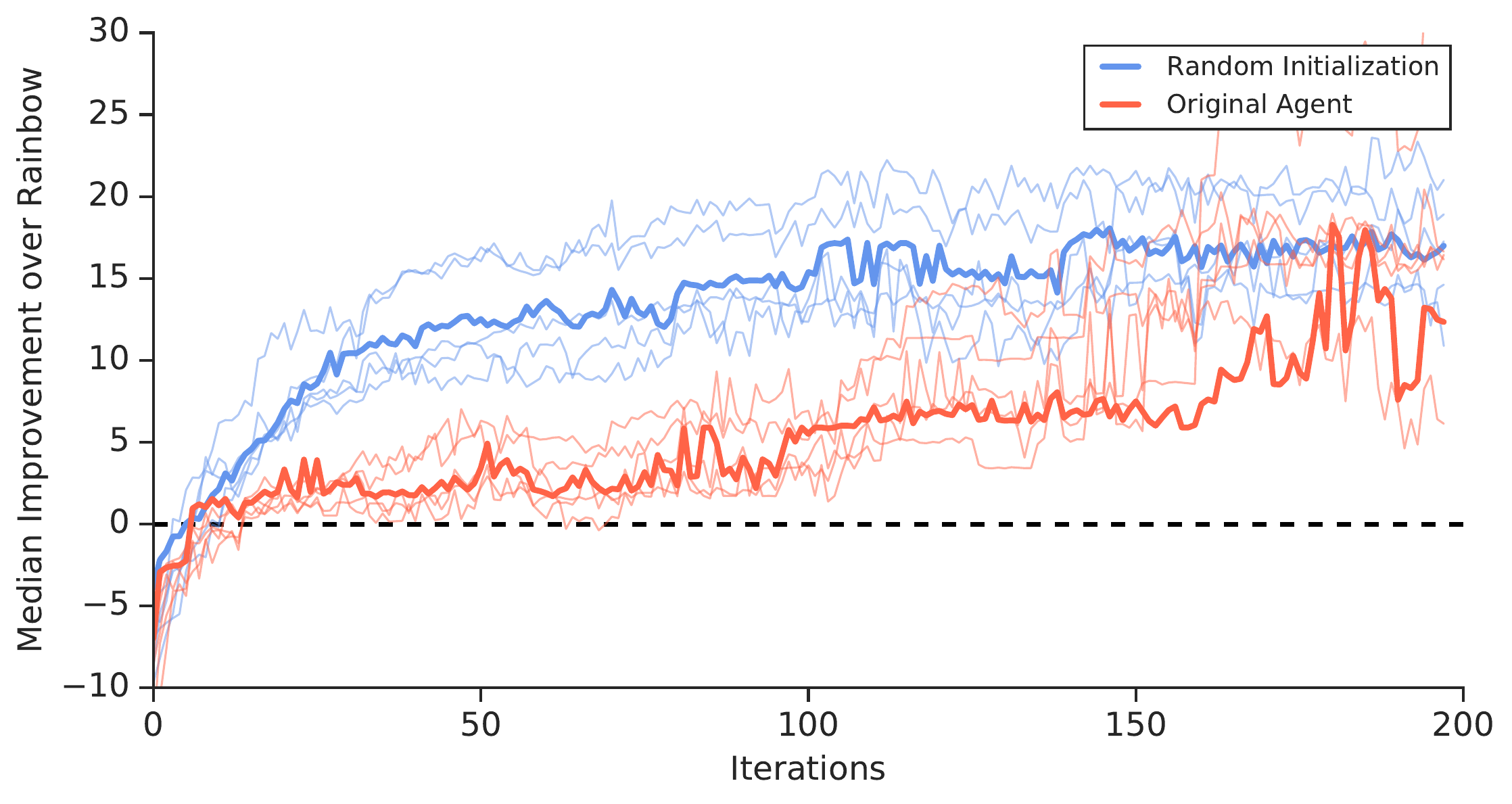}}
    \caption{We examine the impact of the initial parameters for the Rainbow Memento agent.  The median improvement over the corresponding original agent is recorded for both random weight initialization (blue) as well as parameters from the original agent (red).}
\end{figure}

Interestingly, we find opposite conclusions for the Rainbow and DQN agents.
The Memento Rainbow agent benefits from the original agent weights whilst the Memento DQN agent is better off randomly initialized.
We would also highlight that the gap between both variants is not large.
This indicates that the parameters learned from the earlier stages of the game may be of limited benefit.

\clearpage
\section{Memento in Rainbow versus DQN}

\begin{figure}[!ht]
    \centering
    \subfigure[Rainbow Memento agent.]{\includegraphics[height=12cm]{plots/memento-rainbow-single-og_summary.pdf}}
    \subfigure[DQN Memento agent.]{\label{fig: dqn_memento}\includegraphics[height=12cm]{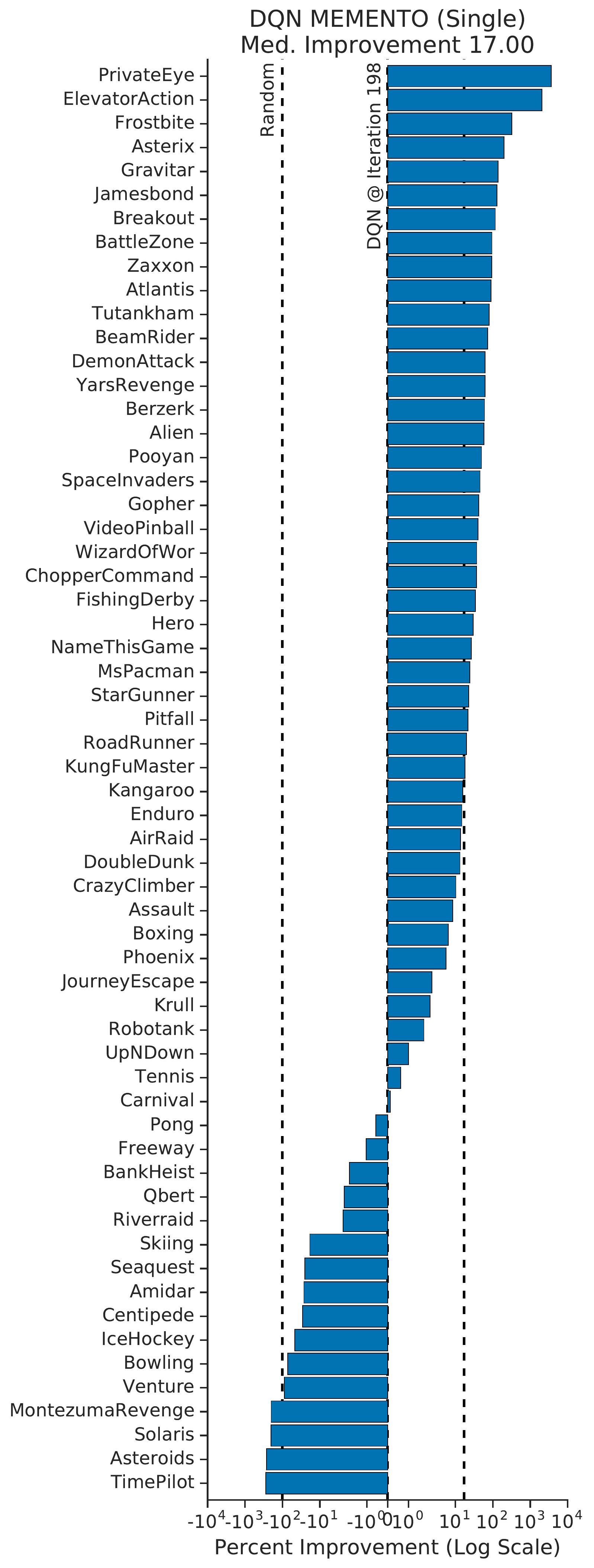}}
    \caption{Each bar corresponds to a different game and the height is the percentage increase of the Memento agent over its baseline.  Across the entire ALE suite, the Rainbow Memento agent (left) improves 75.0\% of the games performance with a median improvement of +25.0\%. The DQN Memento agent (right) improves 73.8\% of the games performance with a median improvement of 17.0\%.}
\end{figure}

\clearpage
\section{Training Curves}
\begin{figure}[!h]
    \centering
    \includegraphics[height=8.6cm,angle=90,origin=c]{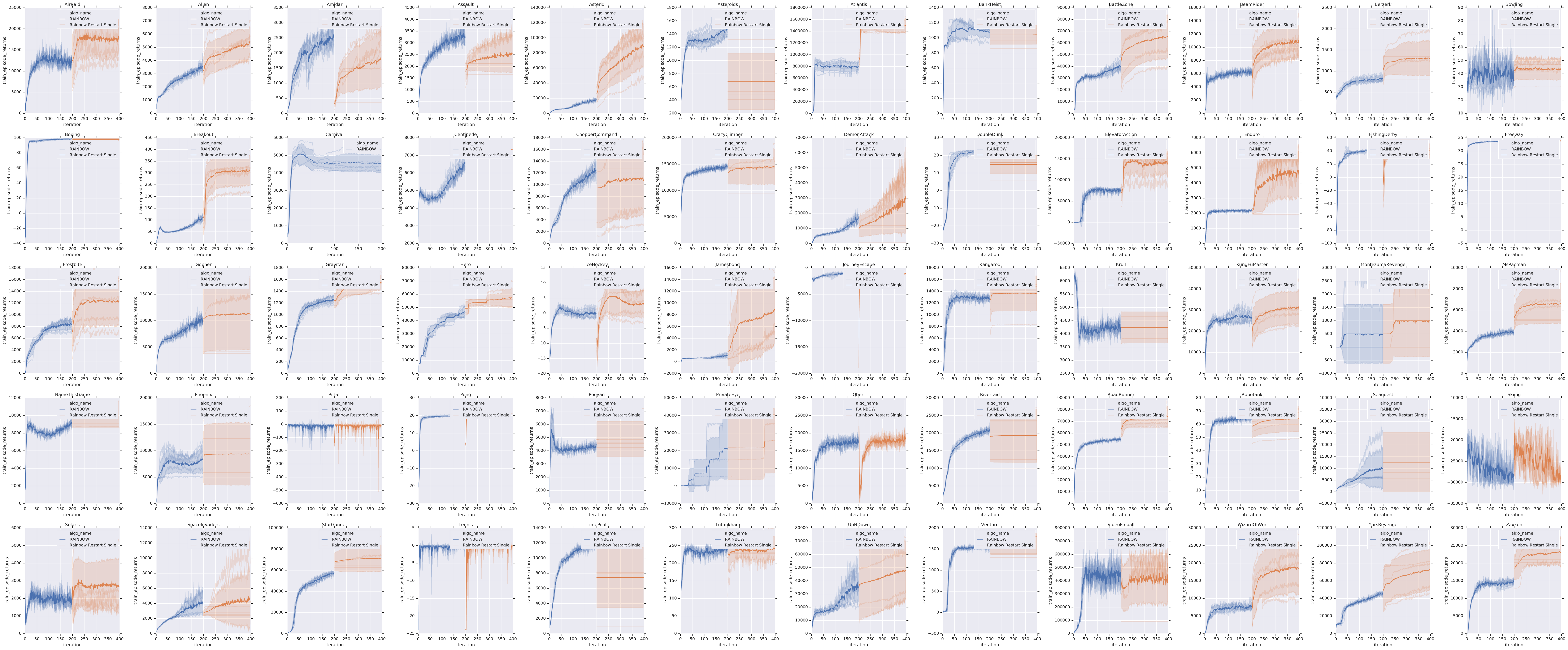}
    \caption{Rainbow MEMENTO training curves.  Blue is the baseline agent and orange is the MEMENTO agent launched from the best position of the previous (five seeds each).}
\end{figure}

\end{document}